\documentclass{llncs}

\usepackage[utf8]{inputenc}
\usepackage{amsmath}
\usepackage{graphicx}
\usepackage{amsfonts}
\usepackage{amssymb}
\usepackage{url}

\usepackage{graphicx}
\usepackage{booktabs}
\usepackage{float}
\usepackage{multirow}
\usepackage{caption}
\usepackage{amsmath}
\usepackage{bm}
\usepackage{cite}
\usepackage{hyperref}

\usepackage{makecell}

\newcommand{\bs}{\boldsymbol}

\begin{document}

{\let\thefootnote\relax\footnotetext{Copyright \textcopyright\ 2020 for this paper by its authors. Use permitted under Creative Commons License Attribution 4.0 International (CC BY 4.0). CLEF 2020, 22-25 September 2020, Thessaloniki, Greece.}}

\title{Priberam at MESINESP Multi-label Classification of Medical Texts Task}

\author{Rúben Cardoso \and
Zita Marinho \and
Afonso Mendes \and
Sebastião Miranda}

\institute{Priberam Labs, Lisbon, Portugal\\
\href{http://labs.priberam.com}{labs.priberam.com}\\
% \url{labs.priberam.com}\\
\email{\{rac,zam,amm,ssm\}@priberam.com}}

\maketitle

\begin{abstract}
Medical articles provide current state of the art treatments and diagnostics to many medical practitioners and professionals. Existing public databases such as MEDLINE contain over 27 million articles, making it difficult to extract relevant content without the use of efficient search engines. Information retrieval tools are crucial in order to navigate and provide meaningful recommendations for articles and treatments. Classifying these articles into broader medical topics can improve the retrieval of related articles~\cite{topic_IR}.
The set of medical labels considered for the \textsc{MESINESP} task is on the order of several thousands of labels (DeCS codes), which falls under the extreme multi-label classification problem~\cite{xml_aggreg}. The heterogeneous and highly hierarchical structure of medical topics makes the task of manually classifying articles extremely laborious and costly. It is, therefore, crucial to automate the process of classification. Typical machine learning algorithms become computationally demanding with such a large number of labels and achieving better recall on such datasets becomes an unsolved problem.

This work presents Priberam's participation at the BioASQ task \textsc{Mesinesp}. We address the large multi-label classification problem through the use of four different models: a Support Vector Machine (SVM)~\cite{svm_liblinear}, a customised search engine (Priberam Search)~\cite{prib_search}, a BERT based classifier~\cite{bert}, and a SVM-rank ensemble~\cite{svm_rank} of all the previous models. Results demonstrate that all three individual models perform well and the best performance is achieved by their ensemble, granting Priberam the $6$\textit{-th} place in the present challenge and making it the $2$\textit{-nd} best team. 
\end{abstract}

\section{Introduction}

A growing number of medical articles is published every year, with a current estimated rate of at least one new article every 26 seconds~\cite{medical_articles}. The large magnitude of both the documents and the assigned topics renders automatic classification algorithms a necessity in organising and providing relevant information. Search engines have a vital role in easing the burden of accessing this information efficiently, however, these usually rely on the manual indexing or tagging of articles, which is a slow and burdensome process~\cite{deep_xml}.

The \textsc{Mesinesp} task consists in automatically indexing abstracts in Spanish from two well-known medical databases, IBECS and LILACS, with tags from a pool of $34118$ hierarchically structured medical terms, the DeCS codes. This trilingual vocabulary (English, Portuguese and Spanish) serves as a unique vocabulary in indexing medical articles. It follows a tree structure that divides the codes into broader classes and more refined sub-classes respecting their conceptual and semantic relationships~\cite{decs_codes}.

In this task, we tackle the \emph{extreme multi-label (XML)} classification problem. Our goal is to predict for a given article the most relevant subset of labels from an extremely large label set (order of tens of thousands) using supervised training.\footnote{The task of multi-label classification differs from multi-class classification in that labels are not exclusive, which enables the assignment of several labels to the same article, making the problem even harder~\cite{dismec}.}
Typical multi-label classification techniques are not suitable for the XML setting, due to its large computational requirements: the large number of labels implies that both label and feature vectors are sparse and exist in high-dimensional spaces; and to address the sparsity of label occurrence, a large number of training instances is required. These factors make the application of such techniques highly demanding in terms of time and memory, increasing the requirements of computational resources. An additional difficulty is related with a large tail of very infrequent labels, making its prediction very hard, due to misclassification of these examples.

The \textsc{Mesinesp} task is even more challenging due to two reasons: first, the articles' labels must be predicted only from the abstracts and titles; and second, all the articles to be classified are in Spanish, which prevents the use of additional resources available only for English, such as BioBERT~\cite{biobert} and ClinicalBERT~\cite{clinicalbert}.

This paper describes our participation at the BioASQ task \textsc{Mesinesp}. We explore the performance of a one-vs.-rest model based on Support Vector Machines (SVM)~\cite{svm_liblinear} as well as that of a proprietary search engine, Priberam Search~\cite{prib_search}, which relies on inverted indexes combined with a k-nearest neighbours classifier. Furthermore, we took advantage of BERT's contextualised embeddings~\cite{bert} and tested three possible classifiers: a linear classifier; a label attention mechanism that leverages label semantics; and a recurrent model that predicts a sequence of labels according to their frequency.
We propose the following contributions:
\begin{itemize}
\item Application of BERT's contextualised embeddings to the task of XML classification, including the exploration of linear, attention based and recurrent classifiers. To the best of our knowledge, this work is the first to apply a pretrained BERT model combined with a recurrent network to the XML classification task.
\item Empirical comparison of a simple one-vs.-rest SVM approach with a more complex model combining a recurrent classifier and BERT embeddings.
\item An ensemble of the previous individual methods using SVM-rank, which was capable of outperforming them.
\end{itemize}
% this paper is organized as follows

\section{Related Work}

Currently, there are two main approaches to XML: embedding based methods and tree based methods.

Embedding based methods deal with the problem of high dimensional feature and label vectors by projecting them onto a lower dimensional space~\cite{deep_xml, otherembed}. During prediction, the compressed representation is projected back onto the space of high dimensional labels. This information bottleneck can often reduce noise and allow for a way of regularising the problem. Although very efficient and fast, this approach assumes that the low-dimensional space is capable of encoding most of the original information. For real world problems, this assumption is often too restrictive and may result in decreased performance.

Tree based approaches intend to learn a hierarchy of features or labels from the training set~\cite{fast_xml, othertree}. Typically, a root node is initialised with the complete set of labels and its children nodes are recursively partitioned until all the leaf nodes contain a small number of labels. During prediction, each article is passed along the tree and the path towards its final leaf node defines the predicted set of labels. These methods tend to be slower than embedding based methods but achieve better performance. However, if a partitioning error is made near the top of the tree, its consequences are propagated to the lower levels.

Furthermore, other methods should be referred due to their simple approach capable of achieving competitive results. Among these, DiSMEC~\cite{dismec} should be highlighted because it follows a one-vs.-rest approach which simply learns a weight vector for each label. The multiplication of such weight vector with the data point feature vector yields a score that allows the classification of the label. Another simple approach consists of performing a set of random projections from the feature space towards a lower dimension space where, for each test data point, a k-nearest neighbours algorithm performs a weighted propagation of the neighbour's labels, based on their similarity~\cite{embarass_xml}.

We propose two new approaches which are substantially distinct from the ones discussed above. The first one uses a search engine based on inverted indexing and the second leverages BERT's contextualised embeddings combined with either a linear or recurrent layer.

% - http://publications.idiap.ch/downloads/papers/2017/Pappas_IJCNLP_2017.pdf
% - http://summa-project.eu/wp-content/uploads/2019/01/Topics.pdf
% - https://arxiv.org/abs/1704.03718 (Deep Extreme Multi-label Learning):
%   |- tree based methods
%   |- embedding based methods
%   |- this article combines both
% - lda, labeled lda (https://arxiv.org/pdf/1709.05480.pdf, https://www-nlp.stanford.edu/cmanning/papers/llda-emnlp09.pdf)
% - Subset Labeled LDA for Large-Scale Multi-Label Classification

\section{XML Classification Models}

 We explore the performance of a one-vs.-rest SVM model in~\S\ref{sec:svm}, and a customised search engine (Priberam Search) in \S\ref{sec:pbasearch}. We further experiment with several classifiers leveraging BERT's contextualised embeddings in \S\ref{sec:bert}. In the end we aggregate the predictions of all of these individual models using a SVM-Rank algorithm in \S\ref{sec:ensemble}.

% \newpage

\subsection{Support Vector Machine}
\label{sec:svm}
Our first baseline consists of a simple Support Vector Machine (SVM) using a one-vs.-rest strategy. We train an independent SVM classifier for each possible label. To reduce the burden of computation we only consider labels with frequency above a given threshold $f_{min}$.
Each classifier weight $\boldsymbol{w}\in\mathbb{R}^{d}$ measures the importance assigned to each feature representation of a given article and it is trained to optimise the max-margin loss of the support vectors and the hyper plane~\cite{svm_liblinear}: % $\boldsymbol{x}_i\in \mathbb{R}^d$~\cite{svm_liblinear}:
\begin{align}\label{svm_eq}
    \min_{\boldsymbol{w}} \frac{1}{2} \boldsymbol{w} \boldsymbol{w}^T + C \sum_{i=1}^l \xi(\boldsymbol{w}; \boldsymbol{x}_i, y_i)\\\notag
    \textnormal{s.t. }y_i(\boldsymbol{w}^{\top}\boldsymbol{x}_i+\boldsymbol{b}\geq 1-\xi_i)
\end{align}
where ($\boldsymbol{x}_i, y_i$) are the article-label pairs, $C$ is the regularisation parameter, $\boldsymbol{b}$ is a bias term and $\xi$ corresponds to a slack function used to penalise incorrectly classified points and $\boldsymbol{w}$ is the vector normal to the decision hyper-plane. We used the abstract's term frequency–inverse document frequency (tf-idf) as features to represent $\boldsymbol{x}_i$.

\subsection{Priberam Search}
\label{sec:pbasearch}
The second model consists of a customised search engine, Priberam Search, based on inverted indexing and retrieval using the Okapi-BM25 algorithm~\cite{prib_search}. 

As depicted in figure \ref{fig:prib_search}, it uses an additional k-nearest neighbours algorithm (k-NN) to obtain the set of $k$ indexed articles closest to a query article in feature space over a database of all articles $\mathcal{A}$. This similarity is based on the frequency of words, lemmas and root-words, as well as label semantics and synonyms. A score is then given to each one of these articles and to each one of their labels and label synonyms, and a weighted sum of these scores yields the final score assigned to each label, as explicit in expression \ref{eq:prib_search}.

\hspace{-0.5cm}
\begin{minipage}{0.45\textwidth}
    \begin{figure}[H]
    \centering
    \includegraphics[width=0.9\textwidth]{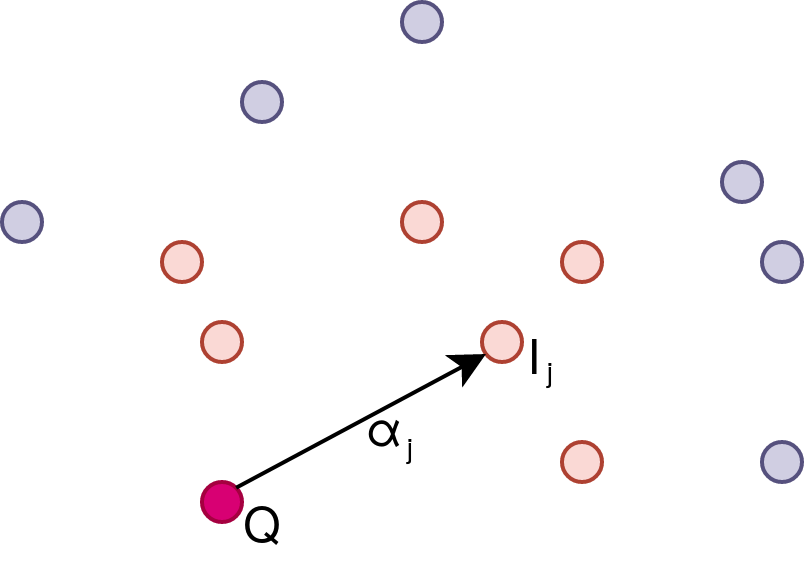}
    \caption{K-NN to obtain $K$ articles $I_{j}\forall j\in[K]$ closest to query article $Q$ in feature space.}\label{fig:prib_search}
    \end{figure}
\end{minipage}
\hspace{1.0cm}
\begin{minipage}{.45\textwidth}
    \begin{equation}\label{eq:prib_search}
        Score_{label\phantom{i}i} = \sum_{j \in \mathcal{A}}^K \alpha_j \cdot \beta_{ji}
    \end{equation}
    \begin{flushleft}
        $\alpha_j$: score of neighbour article $j$ \\
        $\beta_{ji}$: score of label $i$ for article $j$
    \end{flushleft}
\end{minipage}%

% \columnbreak

% \begin{figure}[H]
% \centering
% \includegraphics[width=0.45\textwidth]{knn_draw.png}
% \caption{K-NN to obtain $k$ articles closest to query article in feature space.}
% \end{figure}
% \end{multicols*}

\subsection{XML BERT Classifier}
\label{sec:bert}
Language model pretraining has recently advanced the state of the art in several Natural Language Processing tasks, with the use of contextualised embeddings such as BERT, Bidirectional Encoder Representations from Transformers~\cite{bert}. This model consists of $12$ stacked transformer blocks and its pretraining is performed on a very large corpus following two tasks: next sentence prediction and masked language modelling. The nature of the pretraining tasks makes this model ideal for representing sentence information (given by the representation of the $[CLS]$ token added to the beginning of each sentence). After encoding a sentence with BERT, we apply different classifiers, and fine-tune the model to minimise a multi-label classification loss: 
\begin{align}
\label{eq:bce}
    \textsc{BCELoss}(\bs{x}_i; \bs{y}_i) = y_{i,j} \log \sigma(x_{i,j}) + (1- y_{i,j}) \log (1-\sigma(x_{i,j})),
\end{align}
where $y_{i,j}$ denotes the binary value of label $j$ of article $i$, which is $1$ if it is present and $0$ otherwise, $x_{i,j}$ represents the label predictions (logits) of article $i$ and label $j$, and $\sigma$ is the sigmoid function.

\subsubsection{In-domain transfer knowledge}
Additionally, we performed an extra step of pretraining. Starting from the original weights obtained from BERT pretrained in Spanish, we further pretrained the model with a task of masked language modelling on the corpus composed by all the articles in the training set. This extra step results in more meaningful contextualised representations for this medical corpus, whose domain specific language might differ from the original pretraining corpora. 

After this, we tested three different classifiers: a linear classifier in \S\ref{sec:linearbert}, a linear classifier with label attention in \S\ref{sec:attentionbert} and a recurrent classifier in \S\ref{sec:grubert}.

\subsubsection{XML BERT Linear Classifier}
\label{sec:linearbert}
\begin{figure}[H]
\centering
  \includegraphics[width=0.8\linewidth]{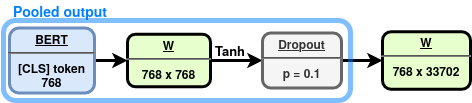}
  \caption{XML BERT Linear Classifier: Flowchart representing BERT's pooled output (in blue) and the simple linear layer (W in green) used as XML classifier.}
  \label{fig:simple_classif}
\end{figure}

The first and simplest classifier consists of a linear layer which maps the sequence output (the $768$ dimensional embedding corresponding to the $[CLS]$ token) to the label space, composed by $33702$ dimensions corresponding to all the labels found in the training set. Such architecture is represented in figure \ref{fig:simple_classif}. We minimise binary cross-entropy using sigmoid activation to allow for multiple active labels per instance, see Eq.~\ref{eq:bce}. This classifier is hereafter designated \textsc{Linear}.

\subsubsection{XML BERT With Label Attention}
\label{sec:attentionbert}
\begin{figure}[H]
  \centering
  \includegraphics[width=0.92\linewidth]{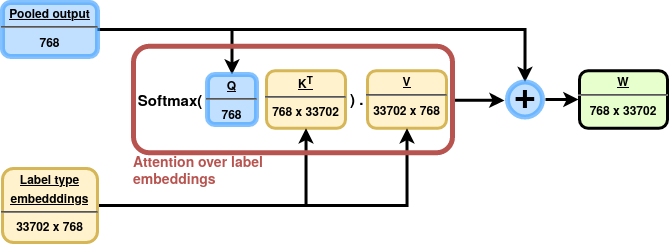}
  \caption{XML BERT with Label Attention Classifier: Article's pooled output (blue) is followed by an extra step of attention over the label embeddings (red) which are finally mapped to a XML linear classifier over labels (green).}
  \label{fig:classif_w_att}
\end{figure}

For the second classifier, we assume a continuous representation with $768$ dimensions for each label. We initialise the label embeddings as the pooled output embeddings (corresponding to the $[CLS]$ token) of a BERT model whose inputs were the string descriptors and synonyms for each label. 
We consider a key-query-value attention mechanism~\cite{att_allneed}, where the query corresponds to the pooled output of the abstract's contextualised representation and the keys and values correspond to the label embeddings. We further consider residual connections, and a final linear layer maps these results to the decision space of $33702$ labels using a linear classifier, as shown in figure \ref{fig:classif_w_att}. Once again, we choose a binary cross-entropy loss (Eq.\ref{eq:bce}). This classifier is hereafter designated \textsc{Label attention}.

\subsubsection{XML BERT With Gated Recurrent Unit}
\label{sec:grubert}
\begin{figure}[H]
  \centering
  \includegraphics[width=1.0\linewidth]{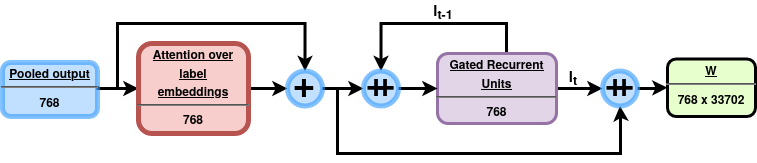}
  \caption{XML BERT GRU Classifier: The GRU network precedes the linear layer and sequentially predicts the labels. The symbol +\!\!\!+ stands for vector concatenation and $l_t$ the label representation predicted by the GRU at time-step $t$.}
  \label{fig:classif_w_gru}
\end{figure}

In the last classifier, we predict the article's labels sequentially. Before the last linear classifier used to project the final representation onto the label space, we add a Gated Recurrent Unit (GRU) network~\cite{gru_article}  with $768$ units that sequentially predicts each label according to label frequency. A flowchart of the architecture is shown in figure \ref{fig:classif_w_gru}. This sequential prediction is performed until the prediction of a stopping label is reached.

We consider a binary cross-entropy loss with two different approaches. On the first approach, all labels are sequentially predicted and the loss is computed only after the stopping label is predicted, i.e., the loss value is independent of the order in which the labels are predicted. It only takes into account the final set. This loss is denominated Bag of Labels loss (BOLL) and it is given by:
\begin{equation}
    \mathcal{L}_{BOLL} = \textsc{BCELoss}(\bs{x}_i; \bs{y}_i)
\end{equation}
where $\bs{x}_i$ and $\bs{y}_i$ are the total set of predicted logits and gold labels for the current article $i$, correspondingly. The models trained with this loss are hereafter designated \textsc{Gru Boll}.

The second approach uses an iterative loss which is computed at each step of the sequential prediction of labels. We compare each predicted label with the gold label, the loss is computed and added to a running loss value. In this case, the loss is denominated Iterative Label loss (ILL):
\begin{equation}
    \mathcal{L}_{ILL} = \sum_{t \in T} \textsc{BCELoss}(\bs x_i^{(t)}; \bs y_i^{(t)})
\end{equation}

where $T$ is the length of the label sequence, $t$ denotes the time-steps taken by the GRU until the ``stop label'' is predicted, and $\bs x_i^{(t)}$ and $\bs y_i^{(t)}$ are the predicted logits and gold labels for time-step $t$ and article $i$, respectively. Models trained with this loss are hereafter designated \textsc{Gru Ill}.

Although only one of the losses accounts directly for prediction order, this factor is always relevant because it affects the final set of predicted labels. This way, the model must be trained and tested assuming a specific label ordering. For this work, we used two orders: ascending and descending label frequency on the training set, designated \textsc{Gru ascend} and \textsc{Gru descend}, respectively.

\begin{table}[H]
\centering
\begin{tabular}{@{}lll@{}}
\toprule
Prediction order \phantom{2222} & Label  & Frequency \\ \midrule \midrule
0 & adulto & 54785 \\ 
1 & niño & 26585 \\ 
... & ... & ... \\ 
13 & prevención de accidentes \phantom{2222} & 349 \\ 
14 & STOP-label & - \\ \bottomrule
\end{tabular}\caption{Example of predictions for a particular abstract using the masking system which enforces a descending label frequency.}\label{tab:gru_example}
\end{table}

Additionally, we developed a masking system to force the sequential prediction of labels according to the chosen frequency order. This means that at each step the output label set is reduced to all labels whose frequency falls bellow or above the previous label, depending on the monotonically ascending or descending order, respectively. Models in which such masking is used are designated \textsc{Gru w/ mask}. Table \ref{tab:gru_example} shows an example of the consecutive predictions obtained for a given abstract throughout the various predictive time-steps.

\subsection{Ensemble}
\label{sec:ensemble}
Furthermore, we developed an ensemble model combining the results of the previously described SVM, Priberam Search and BERT with GRU models. This ensemble's main goal is to leverage the label scores yielded by these three individual models in order to make a more informed decision regarding the relevance of each label to the abstracts.

\begin{figure}[H]
\centering
\includegraphics[width=0.85\textwidth]{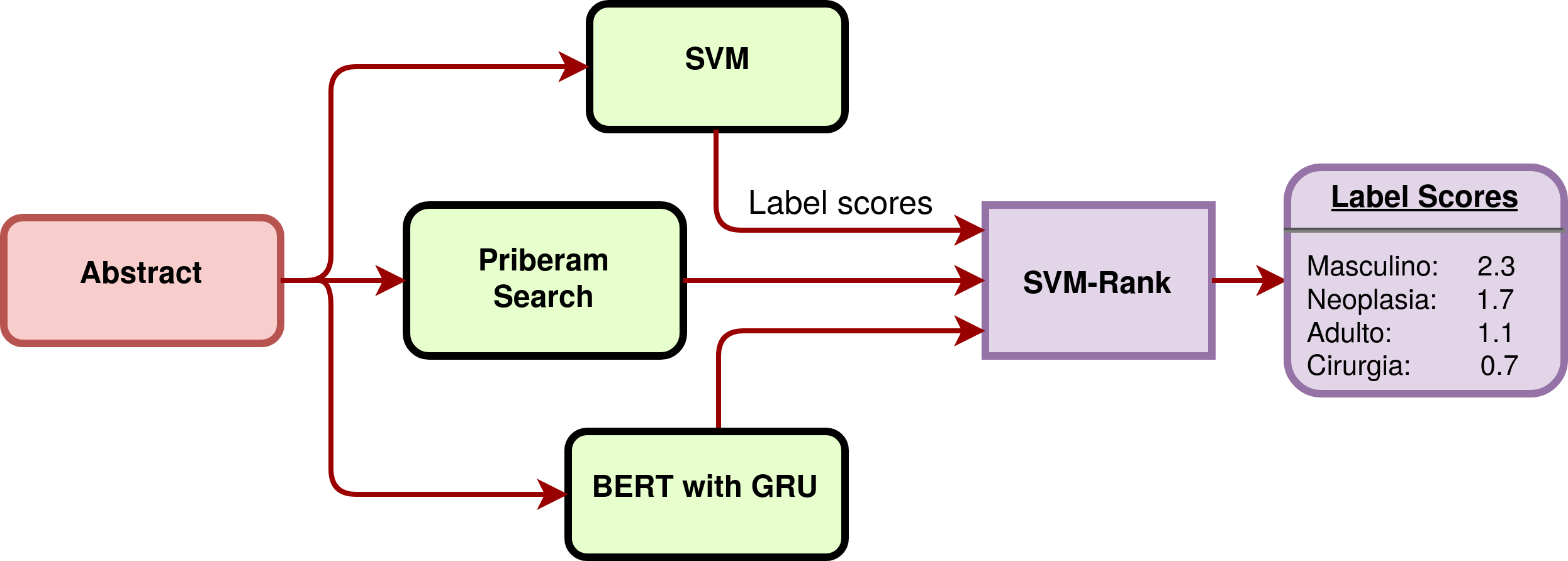}
\caption{SVM-Rank combines the three previous individual models.}\label{fig:ensemble_svmrank}
\end{figure}

We chose an ensembling method based on a SVM-rank algorithm~\cite{svm_rank} whose features are the normalised scores yielded by the three individual models, as well as their pairwise product and full product. These scores are the distance to the hyper-plane in the SVM model, the k-nearest neighbours score for Priberam Search and the label probability for the BERT model. This approach is depicted on figure \ref{fig:ensemble_svmrank}.

An SVM-rank is a variant of the support vector machine algorithm used to solve ranking problems~\cite{learn_to_rank}. It essentially leverages pair-wise ranking methods to sort and score results based on their relevance for a specific query. This algorithm optimises an analogous loss to the one shown in Eq. \ref{svm_eq}. Such ensemble is hereafter designated \textsc{SVM-rank ensemble}.

\section{Experimental Setup}
\label{sec:exp}
We consider the training set provided for the \textsc{Mesinesp} competition containing $318658$ articles with at least one DeCS code and an average of $8.12$ codes per article. We trained the individual models with $95\%$ of this data. The remaining $5\%$ were used to train the SVM-rank algorithm. The provided smaller official development set, with 750 samples, was used to fine-tune the individual model's and ensemble's hyper-parameters, while the test set, with 500 samples, was used for reporting final results. These two sets were manually annotated by experts specifically for the \textsc{MESINESP} task.

\subsection{Support Vector Machine}
For the SVM model we chose to ignore all labels that appeared in less than $20$ abstracts. With this cutoff, we decrease the output label set size to $\approx9200$.
Additionally, we use a linear kernel to reduce computation time and avoid over-fitting, which is critical to train such a large number of classifiers. Regarding regularisation, we obtained the best performance using a regularisation parameter set to $C=1.0$, and a squared hinge slack function whose penalty over the misclassified data points is computed with an $\ell_2$ distance.

Furthermore, to enable more control over the classification boundary, after solving the optimisation problem we moved the decision hyper-plane along the direction of $\boldsymbol{w}$. We empirically determined that a distance of $-0.3$ from its original position resulted in the best $\mu F1$ score. This model was implemented using scikit-learn\footnote{\href{https://scikit-learn.org}{scikit-learn.org}} and the code was made publicly available\footnote{\href{https://github.com/Priberam/mesinesp-svm}{github.com/Priberam/mesinesp-svm}}.

\subsection{Priberam Search}

To use the Priberam Search Engine, we first indexed the training set taking into account the abstract text, title, complete set of gold DeCS codes, and also their corresponding string descriptors along with some synonyms provided\footnote{\url{https://temu.bsc.es/mesinesp/wp-content/uploads/2019/12/DeCS.2019.v5.tsv.zip}}.
We tuned the number of neighbours $k=[10,20,30,40,50,60,70,100,200]$ in the development set for the $k$-NN algorithm and obtained the best results for $k=40$. To decide whether or not a label should be assigned to an article, we fine-tuned a score threshold over the interval $[0.1, 0.5]$ using the official development set, obtaining a best performing value of $0.24$. All labels with score above the threshold were picked as correct labels.

% \newpage

\subsection{BERT}

For all types of BERT classifiers, we used the Transformers and PyTorch Python packages~\cite{transformers, pytorch}.

We initialised BERT's weights from its cased version pretrained on Spanish corpora, bert-base-spanish-wwm-cased\footnote{\url{https://github.com/dccuchile/beto}}.

We further performed a pretraining step on the \textsc{Mesinesp} dataset to obtain better in-domain embeddings. For the pretraining and classification task, table \ref{tab:train_hp} shows the training hyper-parameters.

For all the experiments with BERT, the complete set of DeCS codes was considered as the label set.

\begin{table}[H]
\centering
\begin{tabular}{@{}lcc@{}}
\toprule
Hyper-parameter & Pretraining & Classification \\ \midrule \midrule
Batch size & 4 & 8 \\
Learning rate & $5 \cdot 10^{-5}$ & $2 \cdot 10^{-5}$ \\
Warmup steps & 0 & 4000 \\
Max seq lenght & 512 & 512 \\
Learning rate decay & - & linear \\
Dropout probability & 0.1 & 0.1 \\ \bottomrule
\end{tabular}
\captionof{table}{Training hyper-parameters used for BERT's pretraining and classification tasks.}\label{tab:train_hp}
\end{table}

\subsection{Ensemble}

Our ensemble model aggregates the prediction of all the individual models and produces a final predicted label set for each abstract. To improve recall we lowered the score thresholds used for each individual model until the value for which the average number of predicted labels per abstract was approximately double the average number of gold labels. This ensured that the SVM-rank algorithm was trained with a balanced set, and it also resulted in a system in which the individual models have very high recall and the ensemble model is responsible for precision.

We trained the SVM-rank model with the $5\%$ hold-out data of the training set. This algorithm returns a score for each label in each abstract, making it necessary to define a threshold for classification. This threshold was fine-tuned over the interval $[-0.5, 0.5]$ using the official \textsc{Mesinesp} development set, yielding a best performing cut-off score of $-0.0233$.

We also fine-tuned the regularisation parameter, $C$. We experimented the values $C=[0.01,0.1,0.5,1,5,10]$ obtaining the best performance for $C=0.1$. The current model was implemented using a Python wrapper for the dlib C++ toolkit~\cite{dlib}.

\section{Results}
\label{sec:results}
Table \ref{tab:all_models} shows the $\mu$-precision, $\mu$-recall and $\mu$-F1 metrics for the best performing models described above, evaluated on both the official development and test sets.

The comparison between the scores obtained for the one-vs.-rest {SVM} and {Priberam Search} models shows that the SVM outperforms the $k$-NN based Priberam Search in terms of $\mu$F1, which is mostly due to its higher recall. Note that, although not ideal for multi-label problems, the one-vs.-rest strategy for the SVM model was able to achieve a relatively good performance, even with a significantly reduced label set.

\setlength{\tabcolsep}{7pt}

\begin{table}[H]
\centering
\begin{tabular}{@{}lcccccc@{}}
\toprule
\multirow{2}{*}{Model} & \multicolumn{3}{c}{Development set} & \multicolumn{3}{c}{Test set} \\ \cmidrule(lr){2-4} \cmidrule(lr){5-7}
 & $\mu$P & $\mu$R & $\mu$F1 & $\mu$P & $\mu$R & $\mu$F1 \\ \midrule \midrule
\textsc{SVM} & 0.4216 & 0.3740 & 0.3964 & 0.4183 & $\boldsymbol{0.3789}$ & 0.3976  \\ \midrule
\textsc{Priberam Search} & 0.4471 & 0.3017 & 0.3603 & 0.4571 & 0.2700 & 0.3395 \\ \midrule
\textsc{Bert-Gru Boll ascend} & 0.4130 &$\boldsymbol{0.3823}$ & 0.3971 & 0.4293 & 0.3314 & 0.3740 \\ \midrule
\textsc{SVM-rank ensemble} & $\boldsymbol{0.5056}$ & 0.3456 & $\boldsymbol{0.4105}$ &  $\boldsymbol{0.5336}$ & 0.3320 & $\boldsymbol{0.4093}$ \\ \bottomrule
\end{tabular}
\captionof{table}{Micro precision ($\mu$P), micro recall ($\mu$R) and micro F1 ($\mu$F1) obtained with the 4 submitted models for both the development and test sets. For each metric, the best performing model is identified in bold.}\label{tab:all_models}
\end{table}

Table \ref{tab:bert} shows the performance of several classifiers used with BERT. Note that, for these models, in order to save time and computational resources some tests were stopped before achieving their maximum performance, allowing nonetheless comparison with other models.

We trained linear classifiers using the BERT model with pretraining on the \textsc{MESINESP} corpus for $660k$ steps ($\approx19$ epochs) and without such pretraining (marked with *). Results show that, even with an under-trained classifier, such pretraining is already advantageous. This pretraining was employed for all models combining BERT embeddings with a GRU classifier.
The label-attentive Bert model (\textsc{Gru Boll} ascend) shows negligible impact on performance when compared with the simple linear classifier (\textsc{Linear}).

We consider three varying architectures of the \textsc{Bert-Gru} model: Bag of Labels loss (\textsc{Boll}) or Iterative Label loss (\textsc{Ill}), ascending or descending label frequency, and usage or not of masking. Taking into account the best score achieved, the BOLL loss performs better than the ILL loss, even with a smaller number of training steps. For this BOLL loss, it is also evident that the ordering of labels with ascending frequency outperforms the opposite order, and that masking results in decreased performance.

On the other hand, for the ILL loss, masking improves the achieved score and the ordering of labels with descending frequency shows better results.
The best classifier for a BERT-based model is the GRU network trained with a Bag of Labels loss and with labels provided in ascending frequency order (\textsc{Gru Boll ascend}). This model was further trained for a total of $28$ epochs resulting in a $\mu$F1=$0.4918$ on the $5$\% hold-out of the training set. It is important to notice the performance drop from the $5$\% hold-out data to the official development set. This drop is likely a result of the mismatch between the annotation methods used in the two sets, given that the development set was specifically manually annotated for this task.

\begin{table}[H]
\centering
\begin{tabular}{@{}lccc@{}}
\toprule
BERT classifier & Training steps & $\mu$F1  \\ \toprule \toprule
\textsc{Linear*}  & 220k & 0.4476 \\ \midrule
\textsc{Linear}  & 250k$^\dagger$ & 0.4504  \\ \midrule
\textsc{Label attention*}  & 700k &  0.4460  \\ \midrule
\textsc{Gru Boll ascend}  & 80k &  $\boldsymbol{0.4759}$ \\ \midrule
\textsc{Gru Boll descend} & 40k &   0.4655   \\ \midrule
\textsc{Gru Boll ascend w/ mask} & 100k$^\dagger$ &  0.4352  \\ \midrule
\textsc{Gru Ill descend}  & 240k$^\dagger$ &  0.4258  \\ \midrule
\textsc{Gru Ill descend w/ mask}  & 240k$^\dagger$ &  0.4526 \\ \midrule
\textsc{Gru Ill ascend w/ mask}  & 240k$^\dagger$ &  0.4459  \\ \bottomrule
\end{tabular}
\captionof{table}{$\mu$F1 metric evaluated for the 5\% hold-out of the training set. All models have been pretrained on the \textsc{Mesinesp} corpus, except for those duly marked. BOLL: Bag of Labels loss. ILL: Iterative Label loss. \textbf{*}: not pretrained on \textsc{Mesinesp} corpus. $\boldsymbol{\dagger}$: training stopped before maximum $\mu$F1 was reached.}\label{tab:bert}
\end{table}

Surprisingly, the BERT based model shows worse performance than the SVM on the test set. Despite their very similar $\mu$F1 scores for the development set, the \textsc{BERT-GRU} model suffered a considerable performance drop from the development to the test set due to a decrease in recall. This might indicate some over-fitting of hyper-parameters and a possible mismatch between these two expert annotated sets.

Additionally, as made explicit in table \ref{tab:all_models}, the ensemble combining the results of the SVM, Priberam Search and the best performing BERT based classifier achieved the best performance on the development set, outperforming all the individual models.

Finally, table \ref{tab:postions} shows additional classification metrics for each one of the submitted systems, as well as their rank within the \textsc{Mesinesp} task. The analysis of such results makes clear that for the three considered averages (Micro, Macro and per sample), the SVM model shows the best recall score. For most of the remaining metrics, the SVM-rank ensemble is able to leverage the capabilities of the individual models and achieve considerable performance gains, particularly noticeable for the precision scores.

% Finally, it is interesting to compare the performances obtained for the current \textsc{Mesinesp} task with those of the BioASQ Task A, which correspond essentially to the same multi-label classification problem for two different languages, Spanish and English. The average $\mu$F1 obtained by the $5$ top scoring systems for the \textsc{Mesinesp} task was $0.4189$, while for the last week of BioASQ task A it was $0.6940$. Despite the fact that the total number of labels for the BioASQ task is slightly smaller that those for \textsc{Mesinesp} and the existence of resources for the English language that are not available for Spanish, we argue that these factors are not relevant enough to cause such a notable performance difference. We hypothesise that the mismatch between the train and test sets annotations may be critical in this task.

\begin{table}[]
\centering
\begin{tabular}{cllll}
\hline
Metric & \textsc{SVM} & \thead[l]{\textsc{Priberam} \\ \textsc{Search}} & \thead[l]{\textsc{BERT-GRU}\\ \textsc{boll ascend}} & \thead[l]{\textsc{SVM-rank} \\ \textsc{ensemble}} \\ \hline  \hline
$\mu$F1 & 0.3976 (7º) & 0.3395 (13º) & 0.3740 (9º) & $\boldsymbol{0.4093 (6º)}$ \\ \midrule
$\mu$P & 0.4183 (17º) & 0.4571 (10º) & 0.4293 (15º) & $\boldsymbol{0.5336 (6º)}$ \\ \midrule
$\mu$R & $\boldsymbol{0.3789 (6º)}$ & 0.2700 (13º) & 0.3314 (8º) & 0.3320 (7º) \\ \midrule
MaF1 & $\boldsymbol{0.4183 (8º)}$ & 0.1776 (13º) & 0.2009 (11º) & 0.2115 (10º) \\ \midrule
MaP & 0.4602 (9º) & 0.4971 (8º) & 0.4277 (11º) & $\boldsymbol{0.5944 (3º)}$ \\ \midrule
MaR & $\boldsymbol{0.2609 (8º)}$ & 0.1742 (16º) & 0.2002 (11º) & 0.2024 (10º) \\ \midrule
EbF1 & 0.3976 (7º) & 0.3393 (13º) & 0.3678 (9º) & $\boldsymbol{0.4031 (6º)}$ \\ \midrule
EbP & 0.4451 (15º) & 0.4582 (12º) & 0.4477 (14º) & $\boldsymbol{0.5465 (3º)}$ \\ \midrule
EbR & $\boldsymbol{0.3904 (6º)}$ & 0.2824 (13º) & 0.3463 (8º) & 0.3452 (8º) \\ \bottomrule
\end{tabular}
\captionof{table}{Micro ($\mu$), macro (Ma) and per sample (Eb) averages of the precision, recall and F1 scores, followed by score position within the \textsc{Mesinesp} task. For each metric, the best performing model is  identified in bold.}\label{tab:postions}
\end{table}

\section{Conclusions}
This paper introduces three type of extreme multi label classifiers: an SVM, a $k$-NN based search engine and a series of BERT-based classifiers. Our one-vs.-rest SVM model shows the best performance on all recall metrics. We further provide an empirical comparison of different variants of multi-label BERT-based classifiers, where the Gated Recurrent Unit network with the Bag of Labels loss shows the best results. This model yields slightly better results than the SVM model on the development set, however, due to a drop in recall, under-performs it on the test set.
Finally, the SVM-rank ensemble is able to leverage the label scores yielded by the three individual models and combine them into a final ranking model with a precision gain on all metrics, capable of achieving the highest $\mu$F1 score (being the $6$-th best model in the task).

\section{Acknowledgements}

This work is supported by the Lisbon Regional Operational Programme (Lisboa 2020), under the Portugal 2020 Partnership Agreement, through the European Regional Development Fund (ERDF), within project TRAINER (Nº 045347).

\bibliographystyle{splncs04nat}

\begin{thebibliography}{8}

\bibitem{topic_IR}
Yi X, Allan J. A comparative study of utilizing topic models for information retrieval. European conference on information retrieval, pp. 29-41. Springer (2009).

\bibitem{xml_aggreg}
Shen Y, Yu HF, Sanghavi S, Dhillon I. Extreme Multi-label Classification from Aggregated Labels. arXiv preprint arXiv:2004.00198 (2020).

\bibitem{svm_liblinear}
Fan RE, Chang KW, Hsieh CJ, Wang XR, Lin CJ. LIBLINEAR: A library for large linear classification. Journal of machine learning research (2008).

\bibitem{prib_search}
Miranda S, Nogueira D, Mendes A, Vlachos A, Secker A, Garrett R, Mitchel J, Marinho Z. Automated Fact Checking in the News Room. In The World Wide Web Conference (2019).

\bibitem{bert}
Devlin J, Chang M, Lee K, Toutanova K. BERT:  pretraining  of Deep Bidirectional Transformers for Language Understanding. Proceedings of the 2019 Conference of the North American Chapter of the Association for Computational Linguistics (2019).

\bibitem{svm_rank}
Joachims T. Optimizing search engines using clickthrough data. InProceedings of the eighth ACM SIGKDD international conference on Knowledge discovery and data mining (2002).

\bibitem{medical_articles}
Garba, S., Ahmed, A., Mai, A., Makama, G. and Odigie, V. Proliferations of scientific medical journals: a burden or a blessing. Oman medical journal, 25(4), p.311 (2010).

\bibitem{deep_xml}
Zhang W, Yan J, Wang X, Zha H. Deep extreme multi-label learning. Proceedings of the 2018 ACM on International Conference on Multimedia Retrieval (2018).

\bibitem{decs_codes}
VHL Network Portal. Red.bvsalud.org. 2020. Decs. [online] Available at: http://red.bvsalud.org/decs/en/about-decs/ (Accessed 2 May 2020).

\bibitem{dismec}
Babbar R, Schölkopf B. DiSMEC: Distributed Sparse Machines for Extreme Multi-label Classification. Proceedings of the Tenth ACM International Conference on Web Search and Data Mining (2017).

\bibitem{biobert}
Lee J, Yoon W, Kim S, Kim D, Kim S, So CH, Kang J. BioBERT: a pre-trained biomedical language representation model for biomedical text mining. Bioinformatics (2020 Feb).

\bibitem{clinicalbert}
Alsentzer E, Murphy JR, Boag W, Weng WH, Jin D, Naumann T, McDermott M. Publicly available clinical BERT embeddings. arXiv preprint arXiv:1904.03323 (2019).

\bibitem{otherembed}
Tai F, Lin HT. Multilabel classification with principal label space transformation. Neural Computation (2012).

\bibitem{fast_xml}
Prabhu Y, Varma M. Fastxml: A fast, accurate and stable tree-classifier for extreme multi-label learning. Proceedings of the 20th ACM SIGKDD international conference on Knowledge discovery and data mining (2014).

\bibitem{othertree}
Agrawal R, Gupta A, Prabhu Y, Varma M. Multi-label learning with millions of labels: Recommending advertiser bid phrases for web pages. Proceedings of the 22nd international conference on World Wide Web (2013).

\bibitem{embarass_xml}
Verma Y. An Embarrassingly Simple Baseline for eXtreme Multi-label Prediction. arXiv preprint arXiv:1912.08140 (2019).

\bibitem{att_allneed}
Vaswani A, Shazeer N, Parmar N, Uszkoreit J, Jones L, Gomez AN, Kaiser Ł, Polosukhin I. Attention is all you need. Advances in neural information processing systems, pp. 5998-6008 (2017).

\bibitem{gru_article}
Cho K, van Merriënboer B, Gulcehre C, Bahdanau D, Bougares F, Schwenk H, Bengio Y. Learning Phrase Representations using RNN Encoder–Decoder for Statistical Machine Translation. Proceedings of the 2014 Conference on Empirical Methods in Natural Language Processing (2014).

\bibitem{learn_to_rank}
Liu TY. Learning to rank for information retrieval. Springer Science \& Business Media (2011).

\bibitem{transformers}
Wolf T, Debut L, Sanh V, Chaumond J, Delangue C, Moi A, Cistac P, Rault T, Louf R, Funtowicz M, Brew J. HuggingFace's Transformers: State-of-the-art Natural Language Processing. ArXiv (2019).

\bibitem{pytorch}
Paszke A, Gross S, and Massa F, Lerer A, Bradbury J, Chanan G, et al. PyTorch: An Imperative Style, High-Performance Deep Learning Library. Advances in Neural Information Processing Systems 32, p.8024--8035 (2019)

\bibitem{dlib}
King DE. Dlib-ml: A machine learning toolkit. The Journal of Machine Learning Research (2009).

\end{thebibliography}
%

\end{document}